\documentclass{article}

\usepackage{booktabs}
\usepackage{subcaption}
\usepackage{multirow}
\usepackage{amsmath}
\usepackage{amssymb}
\usepackage{xcolor} 
\usepackage{colortbl} 
\usepackage{graphicx}
\usepackage{amsmath} 

\usepackage{hyperref}
\usepackage{cleveref}
\usepackage{wrapfig}

\definecolor{ourcolor}{RGB}{173, 216, 230} 

 \usepackage[preprint]{corl_2026} 

\title{Think When It Matters: Conditional VLM Reasoning for Social Navigation with RL Policies}

\author{
  Ali Ahmadi, Hamed Rahimi, Adrien Jacquet Crétides,\\ \textbf{Marie Samson, Mahdi Khoramshahi, Mohamed Chetouani}\\
  Institut des Systèmes Intelligents et de Robotique (ISIR)\\
  Sorbonne Université 
  France\\
  \texttt{\{lastname\}@isir.upmc.fr} \\
}

\begin{document}
\maketitle


\begin{abstract}

As mobile robots become more integrated into everyday human environments, social robot navigation is becoming essential for ensuring human comfort, safety, and trust. While reinforcement learning (RL) navigation policies provide the fast inference and reactive behavior necessary for real-time deployment, they still lack flexible semantic reasoning capabilities and often fail to generalize to complex social scenarios. Recent approaches have increasingly turned to vision-language models (VLMs) in place of RL policies to improve semantic and social reasoning in robot navigation. Nevertheless, their high computational cost and slow inference remain major barriers to real-time deployment.
To overcome these limitations, we introduce HUMA (Hybrid Understanding for Multi-modal social Navigation), a hybrid architecture that dynamically balances the computational efficiency of RL policies with the deep semantic understanding of VLMs. Our approach uses a reactive RL policy to handle low-density, routine navigation tasks, while conditioning it on a post-trained high-level VLM when a human enters sensitive situations, such as the robot’s proximity zone. 
We evaluate HUMA on the Social-MP3D and Social-HM3D benchmarks, where it achieves task success improvements of 20\% and 3\%, respectively, while significantly reducing personal space violations and human collisions against state-of-the-art baselines. Extensive ablation studies validate each architectural component, and real-world deployment on the Mirokaï mobile robot 
further demonstrates the practical viability of our approach.

\end{abstract}



\section{Introduction}



As mobile robots increasingly transition from isolated settings into dynamic, human-centric environments such as airports or hospitals, social robot navigation appears as one of the main challenges of Human-Robot Interactions (HRI) in modern societies~\cite{singamaneni2024survey,Triebel2016}.  
Social Navigation (also referred to as Human-aware or Socially-aware Navigation) lies at the intersection of Human-Robot Interaction (HRI) and Robot Motion Planning \cite{KRUSE20131726, singamaneni2024survey}. It concerns the ability of robots to navigate safely and efficiently in human-populated environments while respecting social norm and human safety such as judging a path's clearance relative to human movements and obstacles \cite{aghzal2025vlmeval}. This requires an agent to dynamically read social cues and anticipate human intent, transforming classical Motion Planning into a socially-aware optimization task. By adapting its trajectory to maintain a non-disruptive presence, a socially aware robot allows for safe, intuitive and comfortable interactions in shared spaces \cite{nogueira2025legibility}. 
\begin{figure}
    \centering
    \includegraphics[width=0.8\linewidth]{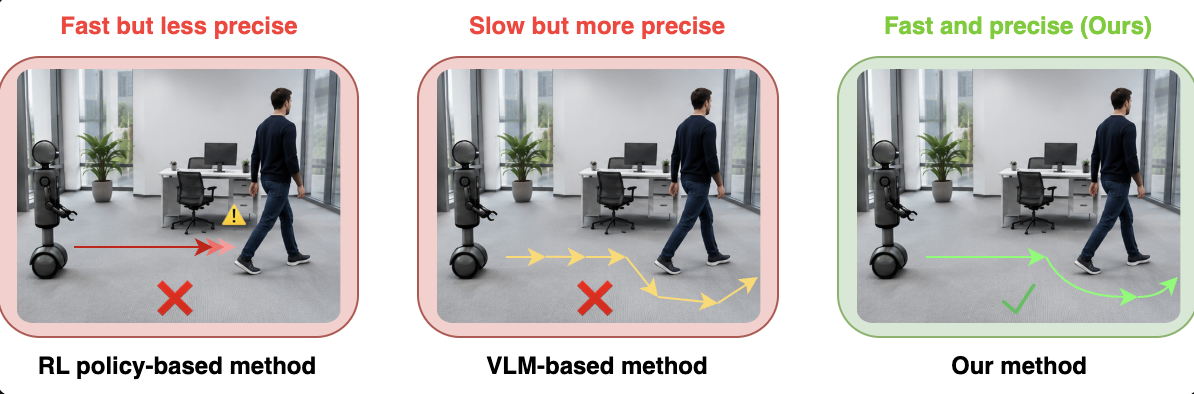}
    \caption{Overview. Existing approaches for social navigation face a fundamental trade-off: RL-based policies (left) offer fast, reactive inference but lack semantic reasoning for complex social scenarios, while VLM-based methods (middle) provide rich contextual understanding at the cost of high computational overhead and slow inference, preventing real-time deployment. Our method (right), combines a reactive RL policy, that handles routine navigation, with a post-trained VLM selectively invoked when necessary, enabling socially compliant, real-time navigation.}
    \label{fig:overview}
\end{figure}
While early frameworks relied on fixed geometric rules, current approaches to social navigation have leveraged learning-based paradigms to tackle complex environments \cite{alyassi2025social,Ramirez2016}. Specifically, policies trained with reinforcement learning (RL) have become a standard approach, largely due to their fast inference and reactive performance in real-time deployment. However, although RL methods are often computationally intensive and data-hungry during training, modeling the nuances of human social behavior typically requires large-scale datasets, and purely RL-based approaches frequently struggle to generalize to rare or edge-case human interactions, resulting in behavior that can be rigid or unnatural.
With recent advances in foundation models, Vision-Language Models (VLMs) have been increasingly incorporated into social robot navigation frameworks due to their strong semantic understanding of visually rich and socially complex environments. VLMs leverages spatial and behavioral reasoning capabilities to decode social contexts, such as pedestrian orientation, group dynamics, or intent, that cannot be captured effectively by RL observations. In spite of their abilities, their massive computational overhead and slow inference time makes their deployment limited for real time navigation scenarios~\cite{Pan2026foundation,huang2024drivlme}. Consequently, the literature lacks a balanced approach that can leverage the deep context-awareness of VLMs without sacrificing the rapid execution speed required for fluid mobile navigation. Our core insight is that high-level semantic reasoning is an unnecessary overhead in classical, uncrowded environments, but becomes critical when maneuvering in proximity to humans. We believe this insight is addressable by employing a adaptive arbitration approach, using a RL policy to handle standard, low-density navigation tasks, while a VLM is activated as the planner the moment a human enters the agent's proximity zone. This allows the robot to interpret complex social contexts and behaviors to guide the underlying navigation controller.

\begin{figure}
    \centering
    \includegraphics[width=0.8\linewidth]{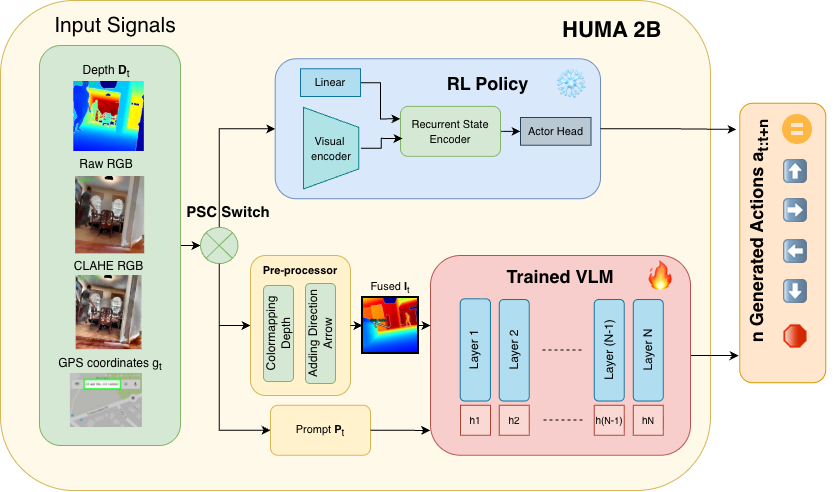}
    \caption{HUMA is a hybrid human-aware navigation framework that dynamically balances computational efficiency with deep semantic understanding. Input signals are routed via a PSC switch to either a lightweight RL policy for standard navigation, or a post-trained VLM pipeline with visual input pre-processing for socially sensitive conditions such as close-proximity human interactions.}
    \label{fig:arch}
\end{figure}

As shown in \Cref{fig:arch}, this paper introduces HUMA, a hybrid human-aware navigation framework that dynamically balances computational efficiency with deep semantic understanding. The HUMA architecture consists of an efficient RL policy for standard navigation, conditioned on a Personal Space Compliance (PSC) switch that enables a VLM for social navigation (trained by LoRA adapters) to intervene under socially sensitive conditions, such as close-proximity human interactions. Through extensive evaluation on the Social-HM3D and Social-MP3D benchmarks \cite{gong2025falcon}, we demonstrate that HUMA significantly outperforms existing baselines, achieving higher social acceptance and improved navigation success rates. Moreover, we perform an ablation study to investigate different configurations of HUMA, and deploy it on the Mirokai robot from Enchanted Tools to showcase its efficiency in real-world scenarios.



\section{Related Work}

\paragraph{Motion Planning in Social Navigation}
Motion Planning is a fundamental challenge in Social Navigation~\cite{KRUSE20131726}, requiring robots to navigate safely, efficiently, and naturally in human-populated environments while adhering to social norms and interaction dynamics. Beyond basic collision avoidance, socially aware navigation demands behaviours that account for human comfort and predictability, such as respecting personal space, adapting motion and speed to surrounding pedestrians, and communicating navigation intent in a socially compliant manner~\cite{singamaneni2024survey}. 
Approaches for this high-level decision-making task mainly consist of planning-based approaches (search and sampling)~\cite{korkmaz2021human,banisetty2021socially,singamaneni2021human,Kollmitz,talebpour2016incorporating,vega2017socially} and learning-based approaches~\cite{luber2012socially,perez2018learning,karnan2022socially,brito2021go}. 
Classical planning methods for robot navigation include search-based approaches such as A* and D* and sampling-based methods such as PRM and RRT~\cite{luo2018porca,charalampous2016robot,korkmaz2021human}. Although recent work improves adaptability through replanning and global-local mapping strategies~\cite{peddi2020data,singamaneni2022watch}, these methods still largely depend on fixed geometric heuristics and explicit maps, motivating the development of learning-based approaches.

\paragraph{Learning-based Motion Planning}
Learning-based methods leverage data-driven models such as Deep Reinforcement Learning (DRL), Convolutional Neural Networks (CNNs), and Inverse RL to predict decisions and adapt to complex environments~\cite{vasquez2014inverse,perez2018learning,brito2021go,valiente2022robustness,Ramirez2016}. Early RL-based methods learn navigation policies end-to-end from sensor observations to control commands, enabling fast inference and efficient obstacle avoidance in crowded environments \cite{liu2020robot}. For instance, Gong et al.~\cite{gong2025falcon} proposed Falcon, a future-aware RL framework for social navigation that explicitly predicts human trajectories. While these approaches improve high-level decision-making and navigation planning, they are often domain-specific, require extensive task-dependent training and data, and generalize poorly to unseen social scenarios due to limited semantic understanding and low interpretability.

\paragraph{VLM-based Motion Planning}
Recent advances in socially aware robot navigation have increasingly shifted toward generative approaches~\cite{sun2024trustnavgpt}, particularly VLM~\cite{chen2025lisn, narasimhan2025olivia}. Recent VLM-based frameworks introduce higher-level contextual reasoning and improved social awareness by leveraging multimodal semantic priors learned from large-scale data. Systems such as CoNVOI~\cite{sathyamoorthy2024convoi}, Walk with Me~\cite{zhang2026walk}, VLM-Social-Nav~\cite{song2024vlm}, and VLM-Informed Path Selection~\cite{fang2026obstacles} exploit vision-language reasoning for socially compliant trajectory evaluation, scene understanding, and instruction following, achieving improved human-aware navigation and stronger generalization across diverse environments. 
While these methods significantly improve success rates and human-aware navigation metrics, they remain computationally expensive, require massive training data and long training times, and often suffer from high inference latency, particularly in long-horizon interactive navigation tasks~\cite{huang2024drivlme}.When comparing VLM-based motion planning with DRL-based approaches, each exhibits complementary strengths and weaknesses: DRL methods typically offer lower planning latency, whereas VLM-based methods provide superior semantic understanding, accuracy, and generalizability. This motivates the exploration of a balanced framework that leverages the advantages of both paradigms to improve overall navigation performance and safety.

\section{Method}



\subsection{Problem Formulation}

We formulate Robot Navigation as a Goal-Conditioned Partially Observable Markov Decision Process (GC-POMDP), $\mathcal{M} = (\mathcal{S}, \mathcal{O}, \mathcal{Z}, \mathcal{G}, \mathcal{A}, \mathcal{T})$, where $\mathcal{S}$ is the state space, $\mathcal{A}$ is the action space, a discrete set of high-level navigational commands, and $\mathcal{T}(\cdot | s, a)$ is the state transition function. 
The agent's goal $g \in \mathcal{G}$ represents the relative coordinate vector $(x_g, y_g) \in \mathbb{R}^2$ to the target location.
At each step $t$, the agent processes an observation $o_t \in \mathcal{O}$ from the state $s_t \in \mathcal{S}$ according to the observation function $\mathcal{Z}(o_t | s_t, a_{t-1})$. $o_t$ is composed of a depth map $\mathbf{D}_t$, the relative goal vector $g_t$ to the robot, and the distance to the closest human $d_{\mathcal{H}}$. 
The agent then queries a policy $\pi$ to select an action $\pi(o_t) = a_t$, with $a_{t} \in \mathcal{A}$, and transitions to the next state $s_{t+1} \sim \mathcal{T}(\cdot | s_t, a_t)$.
In the case of Social Robot Navigation, the state space $\mathcal{S}$ and observation space $\mathcal{O}$ are fundamentally partitioned into distinct regimes depending on the presence and proximity of humans. The objective is to find a global policy $\pi^*(a_t | o_t)$ that successfully reaches a terminal goal state $s_t \approx g$ while minimizing collisions and maintaining socially compliant behavior. 

\subsection{HUMA}

We propose HUMA, a hybrid social robot navigation framework that dynamically switches between a computationally efficient RL policy for standard navigation and a VLM post-trained for socially-aware navigation in the presence of humans. The core architecture comprises three main components: the RL navigation policy, the VLM reasoning module, and a human-dependent switching mechanism.

\paragraph{RL Navigation Policy}
We adopt the RL policy architecture introduced by Falcon \cite{gong2025falcon} as our base navigation policy.
%
%
At each timestep $t$, the policy processes egocentric depth images $\textbf{D}_t$ via a visual encoder alongside point-goal coordinate data $g_t$ transformed via a linear encoder. 
A two-layer LSTM recurrent state encoder extracts latent spatio-temporal features to output discrete navigation actions $a_{t} \in \mathcal{A}$.  

The policy is trained via Decentralized Distributed Proximal Policy Optimization (DD-PPO). To enforce adherence to human social norms without global maps or prior path knowledge, the primary policy updates are driven by a composite reward function $R_{socialnav}^{t}$, which explicitly subtracts a multi-component Social Cognition Penalty $R_{scp}^t$ from goal-directed PointNav rewards such that: $R_{socialnav}^{t}=R_{pointnav}^{t}-R_{scp}^{t}$

The goal-directed PointNav component is defined as $R_{pointnav}^{t}=-\beta_{d}\Delta_{d}-r_{slack}+\beta_{succ}\cdot I_{succ}$, which accounts for the change in distance to the target $\Delta_{d}$, a step penalty $r_{slack}$, and a success indicator $I_{succ}$. This is counterbalanced by the Social Cognition Penalty, formulated as $R_{scp}^{t}=r_{coll}+r_{prox}+r_{traj}$. Within this penalty, the collision component $r_{coll}$ imposes discrete weight penalties for physical contact with static obstacles or human agents. The proximity penalty $r_{prox}$ scales exponentially if the robot breaches a $2.0\text{ m}$ safety envelope around any human agent, computed as $r_{prox}=\sum_{i=1}^{N}\beta_{prox}\cdot \exp(-d_{i}^{t})$ if $d_{i}^{t}<2.0\text{ m}$ and $0$ if $d_{i}^{t}\ge2.0\text{ m}$. Here $N$ is the total number of dynamic humans in the environment, $\beta_{prox}$ represents the penalty weight term for human proximity, and $d_{i}^{t}$ denotes the distance between the robot and the $i$-th human at the current timestep $t$. 
Finally, the trajectory obstruction penalty ($r_{traj}$) penalizes the robot if its immediate path intersects within $0.05\text{ m}$ of a human's projected $H$-step future path, applying heavier weight to imminent path conflicts:

$$r_{traj}=\sum_{k=t+1}^{t+H}\sum_{i=1}^{N}\begin{cases}\beta_{traj}\cdot\left(\frac{1}{k-t+1}\right) & \text{if } d_{traj,i}^{k}<0.05\text{ m} \\ 0 & \text{if } d_{traj,i}^{k}\ge0.05\text{ m}\end{cases}$$

where $H$ is the look-ahead horizon for trajectory forecasting, $\beta_{traj}$ is the penalty weight scaling factor for trajectory obstructions, $k$ represents the specific future timestep being evaluated, and $d_{traj,i}^{k}$ represents the distance between the robot's and the $i$-th human's predicted future positions at the $k$-th forward timestep.

Additionally, the training phase utilizes a Spatial-Temporal Precognition Module. Consequently, the total loss function $\mathcal{L}_{total}$ balances primary navigation PPO updates $\mathcal{L}_{main}$ with the auxiliary objectives loss $\mathcal{L}_{aux}$ such that $\mathcal{L}_{total}=\beta_{main}\mathcal{L}_{main}+\beta_{aux}\mathcal{L}_{aux}$.
$\mathcal{L}_{aux}$ is defined as $\mathcal{L}_{aux}=\mathcal{L}_{count}+\mathcal{L}_{pos}+\mathcal{L}_{traj}$, where $\mathcal{L}_{count}$ is optimized using cross-entropy loss to estimate the total human count , while mean squared error (MSE) is applied to both the current human position tracking $\mathcal{L}_{pos}$ and the multi-step future trajectory forecasting $\mathcal{L}_{traj}$ objectives.

\paragraph{Vision Language Model Reasoning Module}
%
In contrast to RL Policy, the input domain for the VLM, denoted as $o_t^{\text{VLM}}$, integrates pre-processed the multimodal data streams into a single composite image $\mathbf{I}_t^{\text{fused}}$ and a text prompt $\mathbf{P}_t$: $ o_t^{\text{VLM}} = \left( \mathbf{I}_t^{\text{fused}}, \mathbf{P}_t \right)$. To optimize downstream context awareness for the model, we perform a visual token pre-processing step. Raw depth map inputs $\mathbf{D}_t$ are converted into information-dense visual tokens before being fed to the VLM. The depth map is normalized to an 8-bit unsigned integer space, mapped to a 3-channel JET colormap to partition distance zones (close, medium, far), and processed via a Canny edge detector to map structural boundaries. Finally, these binary edges are superimposed as a white contour mask over the colormap alongside the rendered GPS goal vector $g_t$ to create the unified image $\mathbf{I}_t^{\text{fused}}$. In parallel, the GPS goal state is encoded linguistically, with the text prompt $\mathbf{P}_t$ being updated at each timestep $t$ to append state-dependent telemetry to the static social navigation rules: $ \mathbf{P}_t = \left[ \mathbf{P}_{\text{static}}, \mathcal{H}_{\text{text}}(g_t) \right]$ 
where $\mathcal{H}_{\text{text}}$ formats the relative goal coordinates into explicit linguistic descriptions, such as the distance in meters or the target angle in radians. The final inference step passes these inputs to the VLM framework, parameterized by $\theta_{\text{VLM}}$, to generate the next $n$ actions $a_{t:t+n} \in \mathcal{A}$ to perform:

\begin{equation}
    a_{t:t+n} = \pi_\text{VLM}\left( o_t^{VLM}; \theta_{\text{VLM}} \right)
\end{equation}

To bridge the linguistic output of the VLM with the discrete action space $\mathcal{A}$, the model generates specific textual action tokens (e.g., [FORWARD], [LEFT], [STOP]).
To efficiently learn the parameters $\theta_{VLM}$, our model is adapted using Parameter-Efficient Fine-Tuning (PEFT) via Low-Rank Adaptation (LoRA), allowing the model to translate the multimodal image-text input ($o_t^{VLM}$) into a sequence of discrete navigation commands $a_{t:t+n}$.

\paragraph{Switching Mechanism}
%
%
To balance computational efficiency with high-level social reasoning, HUMA employs a dynamic policy switching mechanism.   
Let $d_t = \min_h \|p_{\text{robot}}^t - p_h^t\|_2$ represent the minimum Euclidean distance between the robot and any human at timestep $t$. 
Following Falcon's definition of the PSC Score~\cite{gong2025falcon}, a binary compliance signal is defined based on a $1.0\text{ m}$ threshold (accounting for a $0.3\text{ m}$ human collision radius and $0.25\text{ m}$ robot radius). To prevent recent spatial violations from being smoothed out by long-term metrics, our mechanism monitors a rolling history of the last $W$ binary signals. 
The localized compliance metric, $\text{PSC}_W$, is formulated as:

$$\text{PSC}_W = \frac{1}{W}\sum_{i=t-W+1}^{t} \mathbf{1}[d_i > 1.0\text{ m}]$$

At each timestep, the active policy is selected based on the weighted PSC score $\text{PSC}_W$. The robot follows the efficient RL policy when social compliance remains above a threshold ($\text{PSC}_W \ge \theta$), and switches to the VLM policy when personal space violations occur ($\text{PSC}_W < \theta$) to better handle socially sensitive situations.







\section{Experimental Setup}
\label{sec:setup}

We investigate two main questions in our experiments: (1) whether the high-level social reasoning of the VLM with proximity-based triggering improves performance over recent RL policy baselines, and (2) under which conditions our proposed framework achieves its best performance. To address these questions, we design a controlled experimental setting that systematically evaluates our method against strong RL baselines across varying task configurations and environmental conditions.

\paragraph{Benchmark}
To answer the first research question and evaluate the proposed method in realistic multi-agent settings, we conduct experiments on two photo-realistic simulation benchmarks: \textit{Social-HM3D} and \textit{Social-MP3D}~\cite{gong2025falcon}. \textit{Social-HM3D} comprises 844 unique indoor scenes derived from the Habitat-Matterport 3D (HM3D) dataset~\cite{ramakrishnan2021habitat}.
On the other hand, \textit{Social-MP3D} is built upon the Matterport3D (MP3D) dataset~\cite{chang2017matterport3d} and contains 72 diverse indoor scenes spanning similar semantic categories such as residences, offices, and gyms. 
%
To further investigate the training and evaluation of the VLM component, we additionally evaluate on two social navigation visual question answering datasets: \textit{SNEI} \cite{payandeh2024social} and \textit{MUSON} \cite{liu2025muson}. \textit{SNEI} consists of socially grounded navigation reasoning samples in the form of image-question-answer pairs, constructed on top of the SCAN benchmark~\cite{karnan2022socially}. 
%
%

\paragraph{Metrics} To comprehensively evaluate navigation performance in dynamic human environments, we report a set of complementary metrics that capture success, efficiency, and social compliance. We measure \textit{Success Rate (SR)} as the proportion of episodes in which the agent successfully reaches the goal within a predefined threshold and time limit. To quantify social awareness, we compute \textit{PSC}, defined as the fraction of timesteps during which the robot avoids violating human interpersonal comfort zones, with higher values indicating more socially compliant behavior. Navigation efficiency is assessed using \textit{Success weighted by Path Length (SPL)}~\cite{zhang2025socialnav}, which jointly evaluates success and trajectory optimality relative to the shortest path, where $S_i \in \{0,1\}$ denotes success, $l_i$ is the shortest path length, and $p_i$ is the executed path length. Finally, safety is measured via \textit{Human Collision Rate (H-Coll)}, defined as the percentage of episodes in which the agent collides with at least one pedestrian, where lower values indicate safer navigation behavior.

Additionally, to evaluate the VLM component, we employ \textit{SBERT} and \textit{ROUGE-1} metrics to measure semantic and syntactic alignment between generated and reference responses across multiple social navigation reasoning tasks, including perception and scene understanding, prediction of future human and environmental states, and reasoning and explanation of perceived context, predicted outcomes, and recommended navigation actions in natural language form. Overall, higher values are desirable for SR, PSC, SPL, SBERT, ROUGE-1, while lower values are preferred for H-Coll.


\paragraph{Baselines}
We compare the proposed approach against both classical and learning-based social robot navigation baselines. For classical approaches, we consider \textit{A*}~\cite{hart1968astar} and \textit{ORCA}~\cite{vanderberg2011ocra}. 
For learning-based methods, we evaluate against the official Habitat navigation baseline~\cite{puig2024habitat} and \textit{Falcon}~\cite{gong2025falcon}, which employ RL for embodied navigation tasks in photo-realistic simulation environments. 
Finally, we compare against \textit{NavThinker}~\cite{hu2026navthinker}, a hybrid planning and RL framework that is closely related to our approach in its use of RL as the primary optimization mechanism, while differing substantially in its methodology for imagination and scene understanding. 

\paragraph{Configuration}
For the policy network, we adapt our method on top of the pretrained policy introduced in \textit{Falcon}, where we consider the human collision radius is 0.3m, the robot radius is 0.25m, the PSC distance threshold is set to 1.0m. 
On the other hand, we initialize our VLM using 
Qwen3-VL 2B, a multimodal transformer with approximately 2.1B parameters. The model is fine-tuned using supervised fine-tuning (SFT) with category-specific prompts spanning five task dimensions: perception, prediction, reasoning, action, and explanation. 
To enable efficient adaptation, we employ LoRA within the PEFT framework, restricting updates to a small subset of parameters while preserving the majority of pretrained weights. Specifically, LoRA is applied with rank $r=16$, scaling factor $\alpha=32$, and a dropout rate of 0.05, without bias terms. The adaptation targets the projection and feed-forward modules. This configuration results in approximately 17 million trainable parameters, corresponding to ~0.81\% of the full model. 
We use a per-device batch size of 1 with gradient accumulation over 8 steps, yielding an effective batch size of 8. A warmup ratio of 0.1 is applied to stabilize early-stage optimization. All experiments are conducted on NVIDIA Thor GPUs for both training and evaluation.

\begin{table*}[t]
\centering
\caption{Social navigation results on Social-HM3D and Social-MP3D. Our method achieves the best SR and H-Coll on both benchmarks, with the largest gains on Social-MP3D, while remaining competitive in SPL and PSC. $^\dagger$Results taken from the original paper.}
\label{tab:single_robot_results}
\begin{tabular}{lcccccccc}
\toprule
\multirow{3}{*}{Method}
& \multicolumn{8}{c}{\textbf{Dataset and Metrics}} \\
\cmidrule(lr){2-9}
& \multicolumn{4}{c}{Social-HM3D}
& \multicolumn{4}{c}{Social-MP3D} \\
\cmidrule(lr){2-5}
\cmidrule(lr){6-9}
\multicolumn{1}{c}{}
& SR$\uparrow$
& SPL$\uparrow$
& PSC$\uparrow$
& H-Coll$\downarrow$
& SR$\uparrow$
& SPL$\uparrow$
& PSC$\uparrow$
& H-Coll$\downarrow$ \\
\midrule
A*$^\dagger$ & 44.81 & 43.99 & 90.38 & 54.80 & 45.67 & 44.69 & 91.97 & 54.00 \\
ORCA$^\dagger$ & 37.44 & 32.91 & 92.23 & 39.77 & 38.81 & 34.65 & 94.03 & 39.86 \\
\midrule
Habitat-official$^\dagger$ & 38.99 & 33.53 & 90.37 & 55.48 & 37.00 & 31.76 & 92.03 & 52.33 \\
Falcon$^\dagger$ & 56.26 & 52.05 & 89.76 & 41.22 & 51.67 & 45.54 & 92.53 & 40.67 \\
\midrule
NavThinker$^\dagger$ & 59.46 & \textbf{55.00} & 89.91 & 39.09 & 47.33 & 41.71 & \textbf{93.68} & 37.67 \\
\midrule
\rowcolor{cyan!20}
\textbf{Ours} & \textbf{62.07} & 54.54 & \textbf{92.32} & \textbf{34.48}
& \textbf{70.35} & \textbf{59.41} & 92.96 & \textbf{20.82} \\
\bottomrule
\end{tabular}
\end{table*}

\section{Results}


\paragraph{Comparative Evaluation}
In the first level of evaluation, we compare the proposed method against baseline approaches on the Social-HM3D and Social-MP3D datasets. As reported in \Cref{tab:single_robot_results}, our model achieves an overall accuracy of $62\%$, demonstrating consistent improvements across both benchmarks. In particular, we observe an increase of approximately $20\%$ in success rate on Social-MP3D and $3\%$ on Social-HM3D relative to the baselines, while simultaneously reducing collision rates by $20\%$ and $5\%$, respectively. These results highlight the effectiveness of the proposed approach in improving both task completion and safety. Notably, despite optimizing for higher success rates and lower collisions, the model maintains competitive performance in PSC and SPL metrics, indicating that gains in navigation efficiency and safety do not come at the expense of path quality or adherence to personal space constraints.



\begin{figure}
    \centering
    
    \begin{subfigure}{0.49\linewidth}
        \centering
        \includegraphics[width=\linewidth]{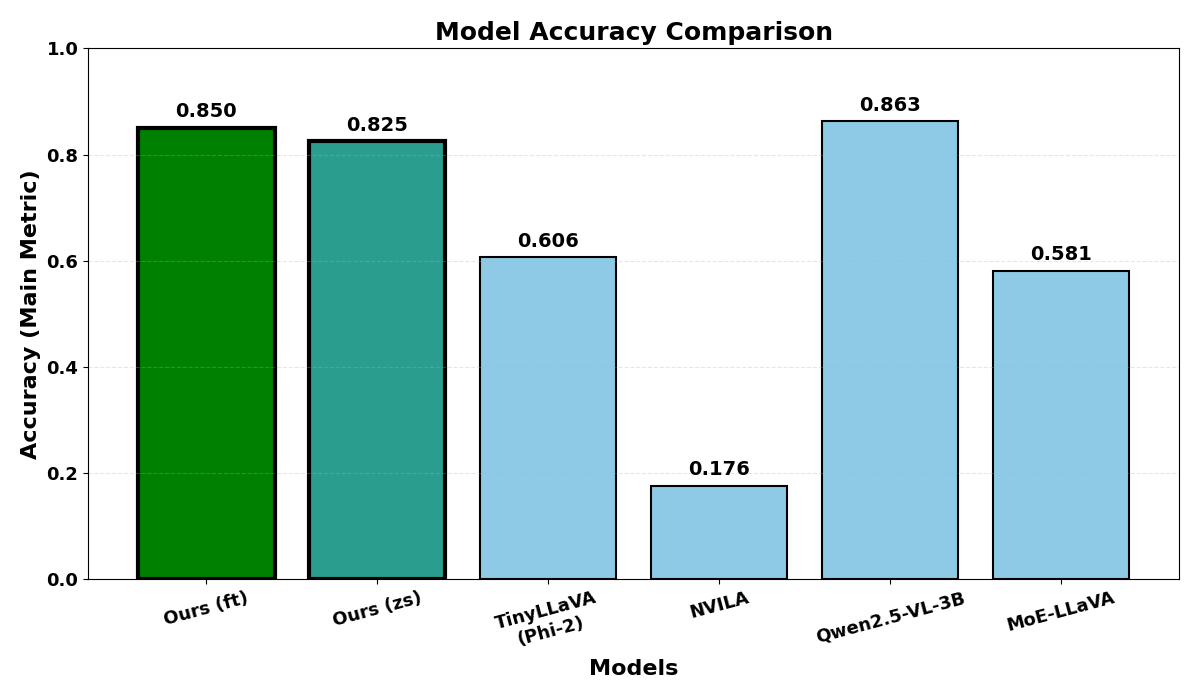}
        \caption{Accuracy comparison across VLM variants}
        \label{fig:vlm_performance}
    \end{subfigure}
    \hfill
    \begin{subfigure}{0.49\linewidth}
        \centering
        \includegraphics[width=\linewidth]{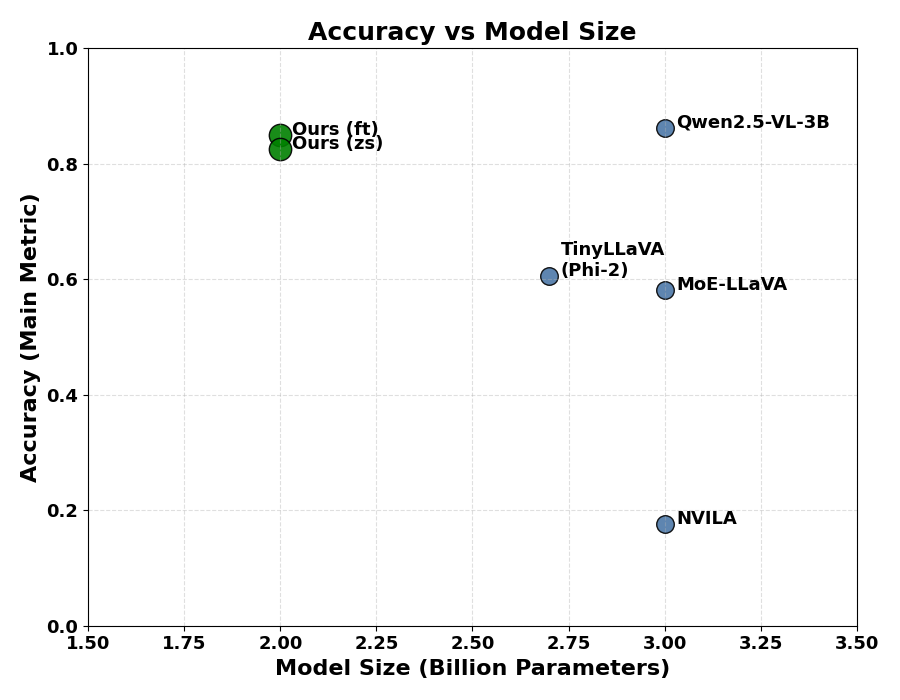}
        \caption{Accuracy vs. model size trade-off}
        \label{fig:vlm_size_ratio}
    \end{subfigure}

    \caption{VLM selection ablation. Our fine-tuned (ft) and zero-shot (zs) 2B models achieve accuracy within 1\% of the best baseline while using ~33\% fewer parameters.}
    \label{fig:vlm_abl_model}
\end{figure}

\paragraph{Ablation Analysis on VLM}
In this ablation study, we investigate the trade-off between inference latency and task performance for the VLM, with the goal of identifying a compact model that maintains strong performance. As shown in \Cref{fig:vlm_performance}, we evaluate multiple model variants and observe that our trained model achieves performance comparable to the highest-performing baseline, with only a $1\%$ difference, while reducing the number of parameters by approximately $33\%$. Furthermore, \Cref{fig:vlm_size_ratio} illustrates the performance-to-parameter ratio, where our approach consistently achieves a significantly higher efficiency compared to competing methods. 
\begin{wraptable}{r}{0.48\textwidth}
\scriptsize
\centering
\caption{Effect of training dataset on VLM performance. Training on SNEI alone yields the best action accuracy, while MUSON improves semantic and syntactic output quality. As action prediction is the primary requirement for our navigation task, SNEI is selected as the training dataset.}
\label{tab:vlm_abl_dataset}
\renewcommand{\arraystretch}{1.2}
\setlength{\tabcolsep}{6pt}
\begin{tabular}{l c c c}
\toprule
Training Dataset & Act. Acc $\uparrow$ & SBERT $\uparrow$ & ROUGE-1 $\uparrow$ \\
\midrule
None (Zero-shot) & 0.825 & 0.507 & 0.317 \\
SNEI & \textbf{0.850} & 0.580 & 0.370 \\
MUSON & 0.762 & \textbf{0.693} & \textbf{0.488} \\
MUSON+SNEI & 0.762 & 0.676 & 0.472 \\
\bottomrule
\end{tabular}
\end{wraptable}
On the other hand, as shown in \Cref{tab:vlm_abl_dataset}, we investigate the impact of different training dataset combinations, specifically SNEI and MUSON, on the performance of our model. Our results indicate that when the Qwen3-VL-2B model is trained on SNEI, it achieves the highest action prediction accuracy on the MUSON evaluation set. In contrast, training on MUSON leads to better semantic and syntactic alignment in the generated reasoning outputs. However, since explicit reasoning generation is not required for our target social navigation task, we prioritize action prediction performance and therefore consider SNEI to be a more suitable training dataset for this setting.

\paragraph{Ablation Analysis on Augmentation of Visual Input}
\label{sec:ablation:input_augmentation}

The visual observations provided by both simulation environments and real-world robotic platforms typically consist of RGB images and depth information. As shown in \Cref{tab:visual_ablation}, we investigate the impact of different input modalities and augmentation strategies on the performance of the proposed framework. In particular, we evaluate standard RGB inputs, colorized depth representations, and their combinations. For the colorized depth setting, depth maps are transformed into colored visual representations accompanied by prompt descriptions, enabling the VLM to better interpret spatial structure and distance information. Furthermore, we analyze the effect of additional visual guidance cues, such as arrows and pointer annotations, by comparing performance with and without these augmentations. Our findings suggest that colorized mapping, together with colored arrow indications for navigation, improves the model’s understanding of the environment, while directional annotations further enhance the reasoning capability of the VLM by providing clearer spatial and navigational context.

\begin{table}[h]
\centering
\caption{Ablation of input modality and visual augmentation. Depth JET colormap yields the best SR, H-Coll, and SPL with the lowest latency, outperforming RGB-based and mixed inputs. Adding directional arrows provides no further gain for depth inputs but degrades performance in RGB settings, confirming colorized depth as the optimal input representation.}
\begin{tabular}{lclllll}
\toprule
\textbf{Type}                          & \textbf{Example}                                                        & \textbf{Setting}         & \textbf{SR$\uparrow$} & \textbf{H-Coll$\downarrow$} & \textbf{SPL$\uparrow$} & \textbf{ms/call} \\ \midrule
\multirow{2}{*}{\small\rotatebox{90}{Depth}} & \multirow{2}{*}{\includegraphics[width=0.75cm]{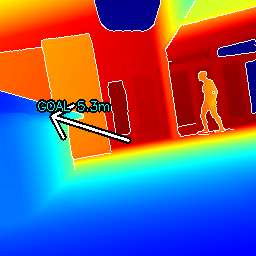 }} & Depth JET (no arrow)     & \textbf{62.07}        & \textbf{34.48}              & \textbf{54.54}         & 2,425            \\
                                       &                                                                         & Depth JET+ green arrow   & \textbf{62.07}        & \textbf{34.48}              & \textbf{54.54}         & 2,425            \\ \midrule
\multirow{2}{*}{\small\rotatebox{90}{RGB}}   & \multirow{2}{*}{\includegraphics[width=0.75cm]{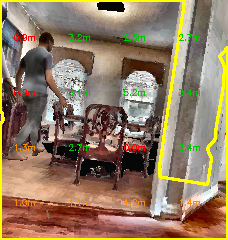}}    & RGB only (CLAHE)         & 59.77                 & 35.63                       & 52.93                  & 4,580            \\
                                       &                                                                         & RGB + white arrow        & 54.02                 & 40.80                       & 48.09                  & 4,260            \\ \midrule
 \multirow{2}{*}{{\small \rotatebox{90}{Mix}}}   & \multirow{2}{*}{\includegraphics[width=1.5cm]{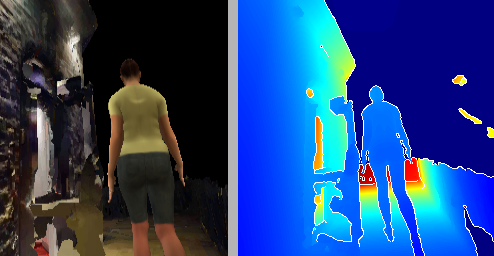}}    & RGB + depth side-by-side & 60.34                 & 35.63                       & 53.41                  & 5,427            \\
                                       &                                                                         & RGB + depth + arrow      & 58.62                 & 38.51                       & 52.93                  & 5,318            \\ \bottomrule
\end{tabular}
\label{tab:visual_ablation}
\end{table}

\paragraph{Ablation Analysis on Proximity Threshold and Evolution of Environment}
\label{sec:ablation_prox_evolution}
We further study the impact of the proximity decision threshold and environment update strategy on navigation performance, as shown in \Cref{tab:nav_results} and \Cref{tab:timing_ablation}. For the proximity threshold, we observe that $r = 0.80$ m yields the best overall performance, achieving the highest success rate and SPL while maintaining the lowest collision rate among all settings. This suggests that an intermediate decision distance provides an effective balance between early intervention and navigation efficiency. 

\begin{table}[h]
    
\centering
\caption{Ablation of PSC switch proximity threshold r. A threshold of 0.80 m achieves the best scores, confirming that intermediate intervention distance optimally trades off safety and navigation efficiency.}
\label{tab:nav_results}
\begin{tabular}{c | c c c c | c}
\toprule
\textbf{$\mathrm{r}$ (m)} & \textbf{SR$\uparrow$} & \textbf{SPL$\uparrow$} & \textbf{PSC$\uparrow$} & \textbf{H-Coll$\downarrow$} & \textbf{\#Calls }\\
\midrule
0.70 & 59.20 & 53.85 & 91.62 & 36.78 & 390 \\
0.75 & 55.75 & 49.93 & 91.75 & 40.23 & 500 \\
0.80 & \textbf{62.07} & \textbf{54.54} & \textbf{92.32} & \textbf{34.48} & 1090 \\
0.85 & 58.05 & 51.90 & 92.20 & 37.93 & 2080 \\
0.90 & 55.75 & 47.66 & 92.72 & 42.53 & 3250 \\
\bottomrule
\end{tabular}
\end{table}

In addition, we analyze different environment evolution strategies during VLM reasoning. As shown in \Cref{tab:timing_ablation}, a fully frozen environment yields the best performance (62.07\% success rate), although this setting is not realistic for real-world deployment. When reducing the frequency of environment updates to more realistic settings, performance slightly degrades from 62\% to 56\% success rate, indicating a trade-off between simulation fidelity and decision stability during VLM inference.


\begin{table}[h]
\centering
\caption{Ablation of environment evolution strategy during VLM inference. Freezing the environment yields the best performance but is unrealistic for deployment; reducing update frequency to more realistic settings incurs a modest 5–6\% SR drop, reflecting a trade-off between decision stability and simulation fidelity.}
\label{tab:timing_ablation}
\setlength{\tabcolsep}{8pt}
\begin{tabular}{l|cccc|c}
\toprule
\textbf{Strategy} & \textbf{SR$\uparrow$} & \textbf{SPL$\uparrow$} & \textbf{PSC$\uparrow$} & \textbf{H-Coll$\downarrow$} & \textbf{\#Calls} \\
\midrule
Freeze & \textbf{62.07} & \textbf{54.54} & \textbf{92.32} & \textbf{34.48} & 109 \\
Pause  & 55.17 & 49.46 & 91.61 & 42.53 & 56 \\
Realistic & 56.90 & 52.31 & 91.62 & 40.23 & 46 \\
\bottomrule
\end{tabular}
\vspace{-10pt}
\end{table}

\begin{figure}[t]
    \centering
    \includegraphics[width=1\linewidth]{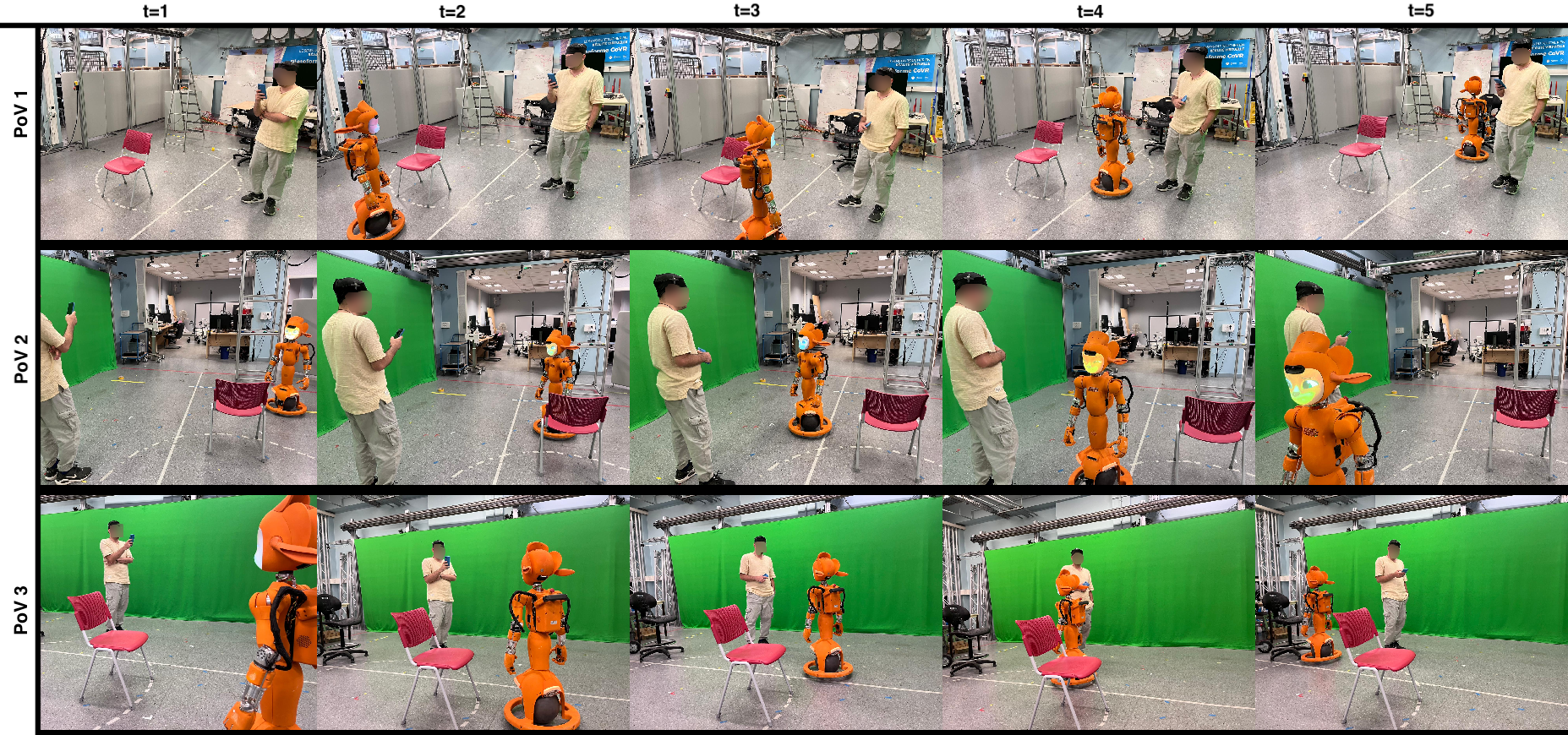}
    \caption{Real-World Deployment of HUMA on the Mirokai robot. The chronological sequence showcases the robot relying on the reactive RL policy to navigate around the chair and reasoning about the human as it gets close to him, before executing a socially compliant trajectory.}
    \label{fig:deployment}
\end{figure}

\section{Deployment}
\label{sec:deploymeny}
To demonstrate HUMA's applicability in a real-world environment, as shown in \Cref{fig:deployment}, we integrated it into the Mirokai humanoid robotics platform developed by \textit{Enchanted Tools}. The deployment architecture bridges real-time data with the robot's base via a ROS2 Humble ecosystem.

\paragraph{Perception Setup} 
Human positions and environment dynamics are captured using a combination of onboard depth camera and an external OptiTrack motion capture system. The motion capture system provides high-precision tracking of agents and goals in the shared space, transformed into the robot's base frame, while the camera data feeds perceptual data to the policy.

\begin{figure}[t]
    \centering
    \includegraphics[width=1\linewidth]{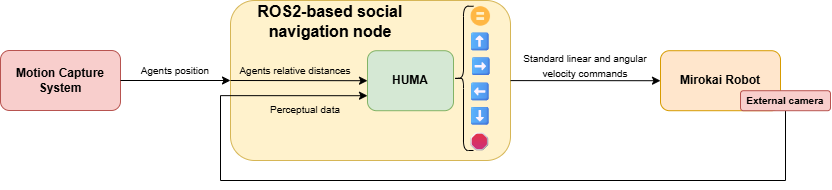}
    \caption{Real-world deployment architecture of HUMA. Agent positions from the OptiTrack motion capture system and perceptual data from the Mirokaï onboard camera are fed as relative distances and raw observations into a ROS2-based social navigation node running HUMA. The framework outputs standard linear and angular velocity commands sent directly to the Mirokaï robot.}
    \label{fig:deploymentpipeline}
\end{figure}

\paragraph{Software Architecture} 
The pipeline, shown in \Cref{fig:deploymentpipeline}, is entirely implemented in ROS2, with dedicated nodes subscribing to the motion capture and camera topics before forwarding the data as input to HUMA. The positions provided by the motion capture system are converted into relative distances between the agents, which are then used as navigation inputs for the policy.

HUMA is encapsulated within a dedicated ROS2 node that processes both the computed distances and the video stream from the onboard camera. The model outputs high-level navigation actions such as \textit{left}, \textit{right}, \textit{forward}, \textit{backward}, \textit{stop}, or \textit{pause}. These discrete actions are then converted within the ROS2 interface node into standard linear and angular velocity commands, which are sent to the robot controller for execution. 

By continuously repeating this process in a closed-loop architecture, the robot is able to navigate toward the target while avoiding surrounding obstacles and humans using HUMA as a ROS2-based social navigation module.

\section*{Limitations}

\begin{wrapfigure}{r}{0.48\textwidth}
    \centering
    \includegraphics[width=\linewidth]{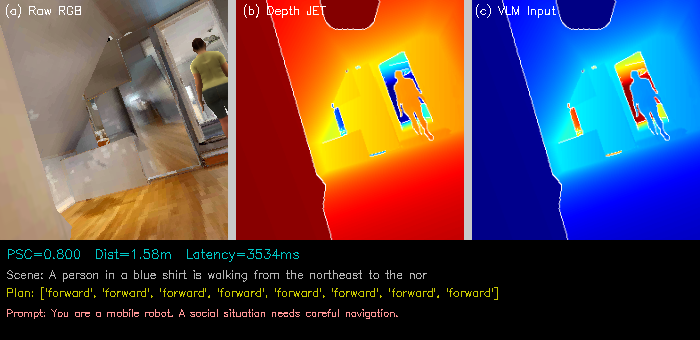}
    \caption{VLM reasoning in a proximity-triggered social navigation scenario. (a) Raw RGB, (b) JET colormap depth, and (c) fused VLM input with Canny edge overlay. With PSC = 0.800 and a human at 1.58 m, the PSC switch activates the VLM, which generates a forward action sequence while describing the social context in natural language. Detailed example provided in \Cref{fig:example_overview}.}
    \label{fig:example_vlm}
\end{wrapfigure}
    
Despite promising results, as an example shown in \Cref{fig:example_vlm}, HUMA has several limitations that motivate future research.  As explained in \Cref{sec:ablation_prox_evolution}, the current freeze timing assumption, where the simulator pauses during VLM inference, does not hold in real deployment, costing $5.2\%$ SR in our realistic timing ablation, and addressing this through better hardware or more efficient models remains an open challenge. Additionally, HUMA reacts to observed PSC violations rather than predicting future ones; integrating Falcon's trajectory prediction module could enable truly proactive intervention. Moreover, as explained in \Cref{sec:ablation_prox_evolution}, the fixed trigger threshold $r= 0.8$ treats all scenes equally, motivating future adaptive mechanisms that adjust based on scene density or pedestrian dynamics. A recurring failure mode is that the VLM correctly handles the immediate social situation but fills remaining plan steps with "forward," suggesting that shorter plans with higher-frequency VLM calls would better support continuous scene reassessment. 
Moreover, as explained in \Cref{sec:ablation:input_augmentation}, the domain gap between the VLM's real-world RGB training data and Habitat's synthetic images also limits performance, particularly in degraded or near-black frames where the model may hallucinate.

\begin{figure}[t]
    \centering
    \includegraphics[width=0.8\linewidth]{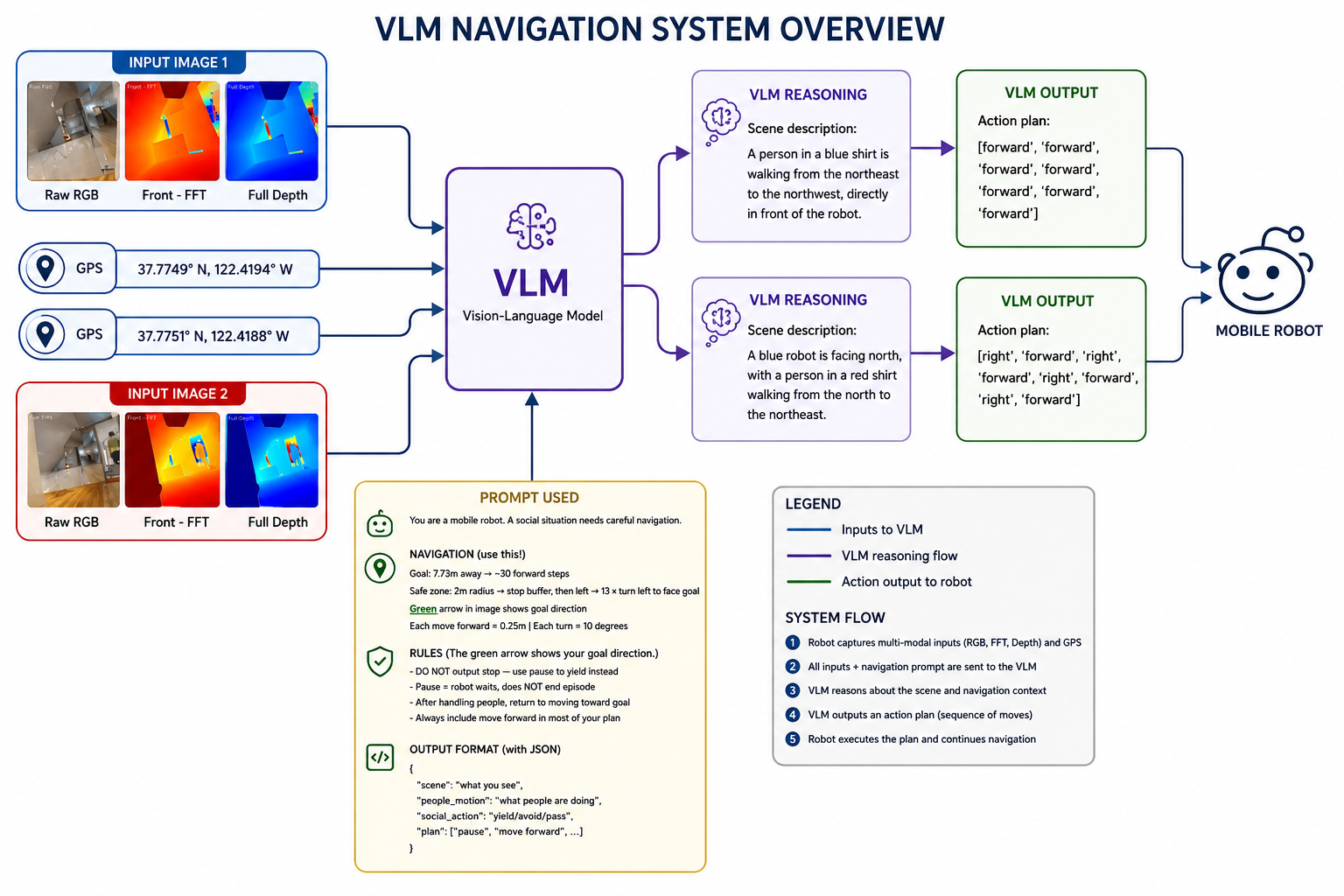}
    \caption{An overview Example. Two consecutive VLM invocations triggered by PSC threshold violations. At each trigger, the robot passes raw RGB, depth JET colormap, and GPS-encoded navigation context to the VLM. The VLM reasons about the social scene and outputs an 8-action plan executed by the robot. RGB and depth images are rendered from the Habitat simulator during evaluation.}
    \label{fig:example_overview}
\end{figure}

To further illustrate the behavior of HUMA during inference, Figure~\ref{fig:example_overview} presents two consecutive VLM invocations triggered by PSC threshold violations. In each case, the raw RGB observation, the JET colormap depth representation, and the GPS-encoded navigation context are passed to the VLM reasoning module. The VLM interprets the social scene and produces an 8-action plan that is subsequently executed by the RL controller. This example highlights two key properties of HUMA: first, the PSC switch correctly identifies socially sensitive situations and activates the VLM only when necessary; second, the fused visual input provides sufficient spatial and semantic context for the VLM to generate socially compliant action sequences. Notably, consecutive invocations demonstrate that the switching mechanism remains responsive across successive timesteps, ensuring continuous social compliance without falling back to the RL policy prematurely.

\section*{Conclusion}
\label{sec:conclusion}
In this paper, we introduced HUMA, a hybrid human-aware navigation framework that dynamically balances the computational efficiency of an RL policy with the deep semantic understanding of a post-trained VLM, activated conditionally via a PSC switch. Evaluated on Social-HM3D and Social-MP3D, HUMA significantly outperforms existing baselines in success rate and human collision rate while maintaining competitive SPL and PSC scores. Extensive ablation studies validate each component of the architecture, and real-world deployment on the Mirokaï robot from Enchanted Tools further demonstrates its practical viability for socially compliant navigation in human-populated environments.


\clearpage


\bibliography{references}  

\clearpage

\appendix


\end{document}